\PassOptionsToPackage{table}{xcolor}
\documentclass[sigconf]{acmart}
\usepackage{multirow}
\usepackage{subfigure}
\usepackage{makecell}
\usepackage{filecontents}
\usepackage{bbm}
\usepackage{pgfplots}
\usetikzlibrary{patterns}
\usepackage[inline,shortlabels]{enumitem}
\usepackage{natbib}
\usepackage{csquotes}
\usepackage{hyperref}
\usepackage{url}
\usepackage{CJKutf8}
\usepackage{algorithm}
\usepackage{svg}
\usepackage{amsmath}
\usepackage{bm}
\usepackage{booktabs} 
\usepackage{graphics}
\usepackage{graphicx}
\usepackage{array}
\usepackage[english]{babel}
\usepackage[T1]{fontenc}
\usepackage{textcomp}
\usepackage{latexsym}
\usepackage[labelformat=simple]{subcaption}
\usepackage{makecell}

\usepackage{siunitx}  
\usepackage[normalem]{ulem}
\useunder{\uline}{\ul}{}
\usepackage{array}

\copyrightyear{2024}
\acmYear{2024}
\setcopyright{acmlicensed}\acmConference[SIGIR '24]{Proceedings of the 47th International ACM SIGIR Conference on Research and Development in Information Retrieval}{July 14--18, 2024}{Washington, DC, USA}
\acmBooktitle{Proceedings of the 47th International ACM SIGIR Conference on Research and Development in Information Retrieval (SIGIR '24), July 14--18, 2024, Washington, DC, USA}
\acmDOI{10.1145/3626772.3657778}
\acmISBN{979-8-4007-0431-4/24/07}

\begin{document}

\title{CorpusLM: Towards a Unified Language Model on Corpus\\ for Knowledge-Intensive Tasks}

\author{Xiaoxi Li}
\affiliation{
 \institution{Gaoling School of Artificial Intelligence}
 \institution{Renmin University of China}
  \city{Beijing}
  \country{China}
}
\email{xiaoxi\_li@ruc.edu.cn}

\author{Zhicheng Dou}
\authornote{Corresponding author.}
\author{Yujia Zhou}
\affiliation{
 \institution{Gaoling School of Artificial Intelligence}
 \institution{Renmin University of China}
  \city{Beijing}
  \country{China}
}
\email{{dou, zhouyujia}@ruc.edu.cn}

\author{Fangchao Liu}
\affiliation{
 \institution{Huawei Poisson Lab}
  \city{Beijing}
  \country{China}
}
\email{liufangchao@huawei.com}

\renewcommand{\shortauthors}{Xiaoxi Li et al.}

\begin{abstract}

Large language models (LLMs) have gained significant attention in various fields but prone to hallucination, especially in knowledge-intensive (KI) tasks. To address this, retrieval-augmented generation (RAG) has emerged as a popular solution to enhance factual accuracy. However, traditional retrieval modules often rely on large document index and disconnect with generative tasks. With the advent of generative retrieval (GR), language models can retrieve by directly generating document identifiers (DocIDs), offering superior performance in retrieval tasks. However, the potential relationship between GR and downstream tasks remains unexplored. 
In this paper, we propose \textbf{CorpusLM}, a unified language model that leverages external corpus to tackle various knowledge-intensive tasks by integrating generative retrieval, closed-book generation, and RAG through a unified greedy decoding process. We design the following mechanisms to facilitate effective retrieval and generation, and improve the end-to-end effectiveness of KI tasks: (1) We develop a ranking-oriented DocID list generation strategy, which refines GR by directly learning from a DocID ranking list, to improve retrieval quality. (2) We design a continuous DocIDs-References-Answer generation strategy, which facilitates effective and efficient RAG. (3) We employ well-designed unsupervised DocID understanding tasks, to comprehend DocID semantics and their relevance to downstream tasks. We evaluate our approach on the widely used KILT benchmark with two variants of backbone models, i.e., T5 and Llama2. Experimental results demonstrate the superior performance of our models in both retrieval and downstream tasks.

\end{abstract}



\maketitle
\title{CorpusLM: Towards a Unified Language Model on Corpus for Knowledge-Intensive Tasks}

\begin{figure*}[t]
    \centering
    \setlength{\abovecaptionskip}{0.14cm}
    \includegraphics[width=0.91\linewidth]{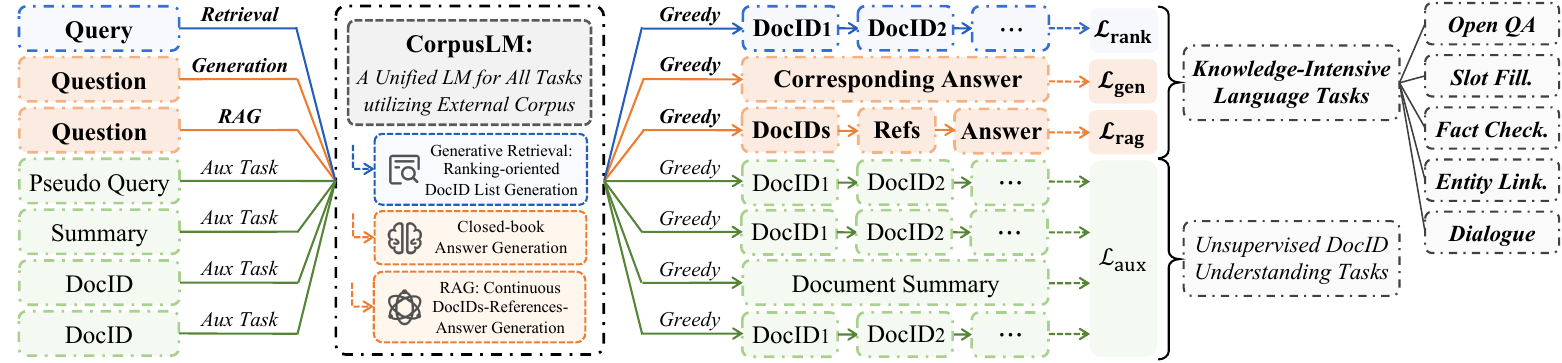}
    \caption{Overview of the CorpusLM framework. 
    We aim to develop a unified language model that utilizes external corpus to handle various knowledge-intensive tasks by integrating generative retrieval, closed-book generation, and RAG. 
    To effectively accomplish all tasks using unified greedy decoding, we propose a ranking-oriented DocID list generation strategy to improve generative retrieval performance; and a continuous RAG strategy to sequentially decode DocID ranking list, references and answer. 
    We also enhance the model's comprehension of DocID semantics through unsupervised DocID understanding tasks.
    }
    \label{fig:model}
\end{figure*}

\section{Introduction}

Large language models (LLMs) have recently revolutionized fields such as question answering (QA), dialogue, and information retrieval, demonstrating impressive capabilities in a variety of language tasks~\cite{gpt3, palm, flan, llama, llm-survey}. However, LLMs often face the problem of ``hallucination'', where the generated text may contain misleading or false information~\cite{hallu-survey}. This issue is particularly severe in knowledge-intensive (KI) tasks, such as slot filling~\cite{trex} and open-domain question answering~\cite{nq}. To address this, a popular approach is the use of retrieval-augmented generation (RAG), which involves a retriever to obtain relevant context from a large knowledge corpus, followed by a generator model that synthesizes the retrieved context into coherent responses~\cite{rag, atlas, llm4ir}.

Traditionally, the retrieval component follows an index-centric framework~\cite{bm25, dpr, ance, colbert, llm-embedder}. Although this approach is widely used, it has several drawbacks. Notably, the requirement for a large document index to search the entire corpus results in considerable memory footprint. Furthermore, during training, the disparate model structures for retrieval (matching similarity) and generation (auto-regressive)~\cite{rag, retro, webgpt, liu2023retallm} hinder the joint optimization of both models. This limitation restricts the understanding of the relationship between both tasks.

Recently, generative retrieval (GR) has emerged as a promising paradigm~\cite{metzler2021rethinking, genre}, which employs auto-regressive generative models to retrieve relevant documents by directly generating document identifiers (DocIDs)~\cite{dsi}. This method has shown improved performance in web search and question-answering (QA) scenarios. While prior research has focused on enhancing model training~\cite{nci, tang2023semantic, zhou2023genrrl}, DocID design~\cite{zhou2023webultron, minder, genret, novo}, and task adaption~\cite{chen2022corpusbrain, chen2023ugr} for better retrieval performance, the unification of generative retriever and downstream generator is often overlooked. Furthermore, the potential of LLMs in the field of generative retrieval remains unexplored.

To address the above problem, we propose \textbf{CorpusLM}, a \textbf{unified} language model that utilizes external corpus and seamlessly integrates generative retrieval, closed-book generation, and retrieval-augmented generation to handle a wide range of knowledge-intensive tasks. These tasks are closely interconnected, necessitating the model to comprehend the relationships between queries, DocIDs, and documents in order to generate accurate answers or DocID ranking lists. To achieve this, we employ a multi-task learning approach, where the generation of both DocIDs and answers is integrated into a unified model training framework. The overview of our framework is illustrated in Figure~\ref{fig:model}.  

Moreover, to facilitate effective retrieval and answer generation in the unified auto-regressive greedy decoding process, we propose the following tailored training and inference strategies: 

\noindent$\bullet$~~\textbf{For retrieval}. Existing methods focus on mapping queries to a single DocID, resulting in high accuracy for the top-ranked document but lower accuracy for subsequent documents~\cite{zhou2023genrrl}. To address this, we introduce a novel \textbf{ranking-oriented DocID list generation} strategy, allowing the model to learn from a ranking list of DocIDs during training. We apply dynamic constraints during the inference to ensure generating a valid and non-repetitive list of DocIDs from the corpus. This strategy offers three advantages: (1)  it enables an end-to-end solution for RAG, as generating a list directly streamlines subsequent steps; (2) training on a query-DocID ranking list can include all relevant DocIDs for a given query, hence it can reduce the training sample confusion problem that exists in existing generative retrieval methods; (3) taking into account the already generated DocIDs during the next DocID decoding can improve the overall ranking capability. 

\noindent$\bullet$~~\textbf{For RAG}, retrieved documents often contain irrelevant and redundant information, hindering the accurate generation of answers. To counteract this, we propose a \textbf{continuous DocIDs-References-Answer generation strategy}. This strategy guides the CorpusLM to first decode DocIDs and their fine-grained references from the documents before decoding the final answer. By doing so, irrelevant information is efficiently filtered out, leading to more accurate responses. Furthermore, our approach enables a continuous decoding process for RAG, eliminating the need for multiple rounds of interaction for document retrieval and answer generation, thereby improving efficiency.

\noindent$\bullet$~~\textbf{DocID understanding}. It is worth noting that existing pre-trained language models do not possess knowledge related to DocIDs. To enhance the model's understanding of DocIDs and establish deeper relationships between DocIDs, queries, and documents, we introduce a set of auxiliary DocID understanding tasks alongside the main tasks of retrieval, generation, and RAG.

We evaluate our model on the KILT benchmark~\cite{kilt}, which is a benchmark for knowledge-intensive language tasks and comprises 11 distinct datasets classified into 5 categories of KI tasks. We implement our model with two different backbone models, which are the encoder-decoder T5 model~\cite{t5} and the open-source decoder-only LLM, Llama2~\cite{llama2}. The experimental results showcase that our proposed CorpusLM achieves superior performance in both retrieval and downstream generation tasks. 






\section{Related Work}

\textbf{Generative Retrieval.}
Generative retrieval (GR) is a novel approach that leverages generative models as differentiable search indices~\cite{dsi}, allowing retrieval by directly generating relevant DocIDs. Recent research in this field mainly focuses on the following aspects:
(1) Model training. A simple yet effective method involves using generated pseudo queries for training data augmentation. Subsequently, labeled query-DocID pairs are used to further fine-tune the model~\cite{dsiqg, nci, zhou2022ultron, tang2023semantic, ltrgr}. To enhance the ranking ability of the GR models,~\citet{zhou2023genrrl} apply reinforcement learning to train a reward model with various annotators, including sparse\&dense retrievers, and LLMs, to provide relevance feedback for GR models.
(2) DocID design. Drawing inspiration from DSI~\cite{dsi}, existing studies explore various approaches such as atomic identifiers, text fragments~\cite{seal, zhou2022ultron, chen2023ugr}, semantic clusters and residual quantization~\cite{mevi}. For instance, Ultron~\cite{zhou2022ultron} utilizes the document URL and title as representations, while SEAL~\cite{seal} considers all n-grams within a corpus as potential identifiers. Recently, researchers have also explored term-sets as DocIDs~\cite{autotsg} and learnable DocIDs~\cite{genret, novo, asi}.
(3) Task adaptation. To better align the GR model with downstream tasks, CorpusBrain~\citet{chen2022corpusbrain} propose pre-training methods that enhance retrieval in knowledge-intensive tasks. Furthermore, UGR~\citet{chen2023ugr} employ n-gram DocIDs tailored for retrieval at various granularities, from sentences to entire documents. However, these methods generally require an additional generator such as FID~\cite{fid} to produce the final output.

\vspace*{1mm}\noindent\textbf{Knowledge-Intensive Language Tasks.}
Knowledge-intensive language tasks refer to a range of NLP tasks that require accessing external knowledge sources to provide accurate results~\cite{fever, nq, hotpotqa, wow, kilt}. These tasks generally involve two components: a retriever and a reader. The retriever obtains relevant information from a large-scale knowledge source and the reader subsequently utilizes specialized downstream models to read the retrieved context and provide more accurate answers~\cite{realm, retro, liu2023retallm, webgpt, webglm}.

To enhance the performance of RAG models, various strategies have been developed, which involve training the retriever and reader modules separately using dense retrieval methods, such as DPR~\cite{dpr}, MT-DPR~\cite{mt-dpr}, DRQA~\cite{drqa} and FID~\cite{fid}. Other strategies focus on joint training of both modules, either by updating the query encoder while keeping the document index static~\cite{rag}, or by updating both the query and document encoders and asymptotically updating the document index during training~\cite{emdr2, sachan2021end, atlas}. 
Based on generative retrieval methods, SEAL~\cite{seal}, CorpusBrain~\cite{chen2022corpusbrain} and UGR~\cite{chen2023ugr} train the GR models and also utilize separate generators to accomplish downstream KILT tasks. Recently, UniGen~\cite{li2023unigen} introduced a unified framework for concurrent learning of GR and QA tasks. However, it uses separate decoders to generate DocIDs and answers, restricting its ability to generalize and scale.

\section{Methodology}

\subsection{Task Formulation}
In this work, we focus on addressing knowledge-intensive language tasks through generative approaches, specifically generative retrieval, closed-book answer generation, and retrieval-augmented answer generation. These tasks can all be accomplished using auto-regressive language models.

Formally, let's consider a document $d$ in a document corpus, and let $d'$ represent the pre-built identifier for document $d$. For \textbf{generative retrieval}, given a query $q$, we determine the relevance $\mathcal{R}$ between $q$ and each document $d$ using the ranking model $f_{\text{rank}}$ with parameters $\theta$:
\begin{equation}
\label{eq:gr}
\mathcal{R}(q,d) = f_{\text{rank}}(d'|q;\theta) = \prod\nolimits_{i=1}^T f_{\text{rank}}(d'_{i}|d'_{<i},q;\theta), 
\end{equation}
where $T$ is the length of the document identifier $d'$, and $d'_i$ refers to the $i$-th token in $d'$, and $d'_{<i}$ represents the generated identifier up to position $i$.

For \textbf{closed-book answer generation}, given an input query $q$, we estimate the probability $\mathcal{A}$ of generating a target answer $a$ using the generation model $f_{\text{gen}}$ with parameters $\phi$:
\begin{equation}
\mathcal{A}_{\text{gen}}(a|q) = f_{\text{gen}}(a|q;\phi) = \prod\nolimits_{i=1}^{T_a} f_{\text{gen}}(a_{i}|a_{<i},q;\phi), \label{eq:gen}
\end{equation}
where $T_a$ is the total number of tokens in $a$, and $a_i$ refers to the $i$-th token in $a$, and $a_{<i}$ represents the generated answer up to index $i$.

For \textbf{retrieval-augmented answer generation}, the probability of generating the target answer $a$ for input query $q$, taking into account the set of retrieved documents $\mathcal{D} = \{d_1, d_2, \dots, d_k\}$, is quantified as follows:
\begin{equation}
\mathcal{A}_{\text{rag}}(a|q,\mathcal{D}) = f_{\text{rag}}(a|q,\mathcal{D};\mu) = \prod\nolimits_{i=1}^{T_a} f_{\text{rag}}(a_{i}|a_{<i},q,\mathcal{D};\mu), \label{eq:rag}
\end{equation}
where $f_{\text{rag}}$ represents the model that generates the final answer after considering the retrieved documents, parameterized by $\mu$.

In this paper, we aim to develop a unified language model for all tasks, so we will use a consistent symbol $f$ with parameters $\theta$ to denote all tasks defined in Equation~(\ref{eq:gr})-(\ref{eq:rag}) in subsequent sections.

\begin{figure*}[!t]
    \centering
    \setlength{\abovecaptionskip}{0.2cm}
    \includegraphics[width=0.85\linewidth]{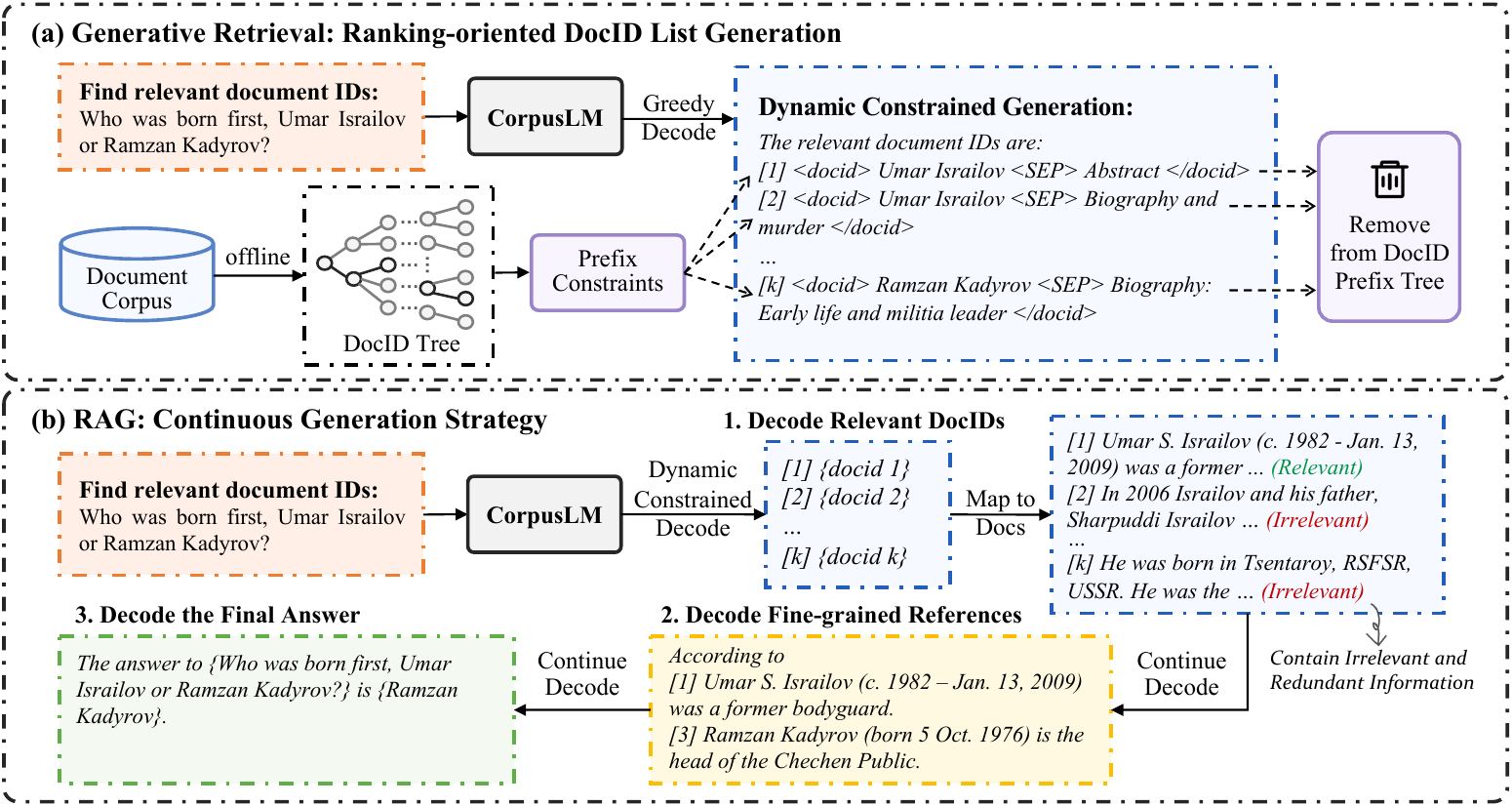}
    \caption{
    Illustration of the two generation strategies for retrieval and RAG. (a) Ranking-oriented DocID list generation strategy. We utilize a prefix tree built from DocIDs in the corpus to dynamically add constraints for generating valid and non-repeated DocID ranking list. (b) Continuous generation strategy. It comprises three continuous decoding steps: (1) decode DocIDs and map them to the corresponding documents; (2) decode fine-grained references from the documents; (3) decode the final answer.
    }
    \label{fig:decodings}
\end{figure*}

\subsection{CorpusLM: the Unified Language Model}
CorpusLM is a multi-task learning architecture designed to handle various types of knowledge-intensive tasks. It is a unified language model capable of performing generative retrieval, closed-book generation, and retrieval-augmented generation through the same auto-regressive greedy generation. The model identifies different tasks using specific prefixes. 

The overview of CorpusLM is illustrated in Figure~\ref{fig:model}. The training of CorpusLM involves the following three basic tasks:
\begin{itemize}[leftmargin=*]
\item \textbf{{Generative Retrieval}:} retrieving relevant documents to a given query by generating a ranked DocID list, facilitating the model's ranking ability, and can be achieved through greedy decoding. 
\item \textbf{{Closed-book Generation}:} generating answers solely based on the query input, without relying on external information, similar to classic auto-regressive language models.
\item \textbf{{Retrieval-Augmented Generation}:} generating answers by first retrieving relevant content using DocID list generation, and then generate references and final response through continuous greedy decoding, enhancing effective and efficient RAG.
\end{itemize}
Moreover, in order to effectively integrate generative retrieval and RAG, it is necessary to improve the model's understanding of DocIDs and the relationship between DocIDs and their corresponding knowledge. Therefore, we incorporate a group of unsupervised DocID understanding tasks into the multi-task learning framework to enhance the model's understanding of the meaning behind DocIDs:
\begin{itemize}[leftmargin=*]
\item \textbf{{DocID Understanding}:} The model is equipped with auxiliary tasks that deepen its understanding of DocIDs' structure and semantic meaning.
\end{itemize}

By training on the above four types of tasks that share common patterns and connections, CorpusLM develops a more comprehensive understanding of the relationships between retrieval and downstream tasks, as well as the meaning behind the DocIDs, thereby gaining a more robust grasp of each individual task. 
%
Formally, the training of CorpusLM aims to optimize objectives with a combined loss function as below:
\begin{equation}
\label{eq:loss_sum}
    \mathcal{L} = \lambda_1 \mathcal{L}_{\text{rank}} + \lambda_2 \mathcal{L}_{\text{gen}} + \lambda_3 \mathcal{L}_{\text{rag}} + \lambda_4 \mathcal{L}_{\text{aux}},
\end{equation}
where $\mathcal{L}{\text{rank}}$, $\mathcal{L}_{\text{gen}}$, and $\mathcal{L}{\text{rag}}$ are corresponding loss functions for generative retrieval, closed-book generation, and retrieval-augmented generation represented by Equations~(\ref{eq:gr})-(\ref{eq:rag}). $\mathcal{L}{\text{aux}}$ is the loss function for the DocID Understanding task. The specific forms of these loss functions will be explained in subsequent sections.
$\lambda_1$, $\lambda_2$, $\lambda_3$, and $\lambda_4$ are weighting coefficients for each task's loss. 

\subsection{Generative Retrieval: DocID List Generation}
To address the limitation of traditional generative retrieval models, which excel at generating the top-1 DocID but often lose accuracy on subsequent DocIDs~\cite{zhou2023genrrl}, we introduce a ranking-oriented DocID decoding strategy in CorpusLM. Instead of using single query-DocID training pairs and beam search, our strategy allows the model to learn from an entire DocID ranking list. By using the greedy decoding method, the model can continuously generate a DocID ranking list using natural language expressions. This approach improves the ranking capability of CorpusLM and promotes joint learning between retrieval and downstream generation tasks.

\subsubsection{Training Data Construction} 
\label{data_construct}
The training process involves learning from a ranked list of relevant DocIDs. However, the challenge lies in the fact that training queries are typically associated with just one or a few relevant documents, which limits the model's capacity to learn effective ranking. 
To overcome this, we enhance the training data with additional relevant documents. We start with the original training queries and their respective answers. We first use a BM25~\cite{bm25} retriever to produce an initial list of potentially relevant documents. Following this, we employ a dense re-ranker SimLM~\cite{simlm} to re-rank the retrieved documents. We create a final list of top-k DocIDs by taking the reranked documents and appending them to the original labeled DocIDs, ensuring there are no duplicates in the combined list. 

Through this method, each query in the training set is associated with a comprehensive list of top-k relevant DocIDs, enabling the model to understand the relevance between multiple ranked DocIDs and the given query, thus reducing training confusion and improving its ranking capability. The final training data format is depicted in Figure~\ref{fig:decodings}, where all of the relevant DocIDs is formatted in a single natural language sequence.

\subsubsection{Training Objective} 
The optimization goal for this task is to maximize the overall relevance of the ranked list of document identifiers \(d'\) for each query \(q\), as defined in Equation~(\ref{eq:gr}). The optimization objective for generating the ranked list of DocIDs is formalized as: 
$\mathcal{O}_{\text{rank}} = \max_{\theta} \sum\nolimits_{d \in \mathbf{D}} \mathcal{R}(q,d)$, where \(\mathbf{D}\) represents the set of documents that are to be ranked for the query \(q\). 

To optimize $\mathcal{O}_{\text{rank}}$ and perform ranking in a natural language format, the ranking loss $\mathcal{L}_{\text{rank}}$ of generative retrieval is defined as:
\begin{equation}
    \mathcal{L}_{\text{rank}} 
    =- \sum\nolimits_{i=1}^{T_d} \text{log} f(s_{i}|s_{<i},q; \theta), \label{eq:loss_rank}
\end{equation}
where $T_d$ denotes the length of tokens in the sequence comprised of the target ranked DocID list. The ranking loss is designed to encourage the model to predict the correct sequence of ranked DocIDs by maximizing the log-likelihood of each token $s_{i}$, given the preceding tokens $s_{<i}$, and the query $q$. 

This approach helps to ensure that the model can generate a sequence of DocIDs that reflects the correct ranking order, also enable ranking-aware generative retrieval in natural language expressions, ultimately improving the retrieval performance of CorpusLM and unification with other generative NLP tasks.

\subsubsection{Inference Constraints} 
During inference, we employ a DocID prefix tree constructed from the documents in the corpus to add constraints that ensure the generation of valid, non-repetitive DocIDs. As depicted in Figure~\ref{fig:decodings}(a), the process involves dynamic activation of constraints during the generation of each token: 

(1) If the token is the DocID start symbol "\texttt{<docid>}", we enforce prefix constraints on the succeeding sequence.

(2) If the token is the DocID end symbol "\texttt{</docid>}", we lift the constraints and remove the previously generated DocID from the prefix tree to prevent duplication in subsequent generations.

(3) For all other tokens, we scan backward; if "\texttt{</docid>}" is encountered, scanning exits. If "\texttt{<docid>}" is found, constraints are applied to the sequence following to ensure the current token's validity within the prefix tree.

\subsection{Closed-book Answer Generation}
For closed-book answer generation, similar to auto-regressive language models like T5 \cite{t5} and GPT \cite{gpt}, the model's objective is to maximize the probability in Equation~(\ref{eq:gen}) with the following loss:
\begin{equation}
    \mathcal{L}_{\text{gen}} = - \sum\nolimits_{i=1}^{T_a} \text{log} f(a_{i}|a_{<i},q;\theta).\label{eq:loss_gen}
\end{equation}

\subsection{RAG: Continuous Generation Strategy}
In order to achieve all KI tasks in a unified greedy decoding process and facilitate effective and efficient retrieval-augmented generation, this section introduces the continuous DocIDs-References-Answer decoding strategy. Notably, in traditional RAG methods, the retrieved documents often contain irrelevant and redundant information that can hinder the accurate generation of answers. Therefore, we propose decoding fine-grained references extracted from the documents first before decoding the final answer.

\textbf{Three-stage Decoding:} As shown in Figure~\ref{fig:decodings}(b), our RAG decoding method consists of three steps:
(1) Decode relevant DocIDs and map them to the corresponding documents in the decoder of CorpusLM.
(2) Continue to decode fine-grained references extracted from the documents.
(3) Continue to decode the final answer.

\textbf{Continuous Decoding:} We achieve these three decoding steps within a single decoding process, rather than reorganizing the input content and relying on multiple input-output iterations. Instead, after retrieving the documents, we directly continue decoding the references and then proceed to decode the final answer.

\textbf{Noise Sampling:} To prevent the model from overly relying on the previous reference when generating the final answer, we introduce the noise sampling strategy. During training, there is a probability \(\tau\) of randomly selecting a sentence from the retrieved documents and replacing the original ground-truth reference.

\textbf{Training Objective:} Given a query \(q\), the model's objective is to first generate a reference \(r\) after retrieving the relevant documents \(\mathcal{D} = \{d_1, d_2, \dots, d_k\}\). The training process involves minimizing the loss function \(\mathcal{L}_{\text{ref}}\), which is defined as:
\begin{equation}
\label{eq:loss_ref}
    \mathcal{L}_{\text{ref}} = - \sum\nolimits_{i=1}^{T_r} \text{log} f(r_{i}|r_{<i},q,\mathcal{D};\theta),
\end{equation}
where \(T_r\) is the length of the reference \(r\), and \(r_{i}\) denotes the \(i\)-th token in the reference sequence. This expects the model to learn to extract fine-grained references after the retrieved passages.

Subsequently, the model generates the final answer \(a\). The loss function for answer generation is defined as:
\begin{equation}
\begin{aligned}
\label{eq:loss_ans}
    \mathcal{L}_{\text{ans}} = &- \sum\nolimits_{i=1}^{T_a} \text{log} f(a_{i}|a_{<i},q,\mathcal{D},r;\theta))\\
    &- \tau\sum\nolimits_{i=1}^{T_a} \text{log} f\left(a_{i}|a_{<i},q,\mathcal{D},r_{\text{noise}};\theta)\right),
\end{aligned}
\end{equation}
where \(T_a\) is the length of the answer \(a\), \(\mathcal{D}\) represents the set of retrieved documents used during training, and $r_{\text{noise}}$ denotes the randomly sampled sentence used as the noised reference.

The overall training loss $\mathcal{L}_{\text{rag}}$ is the sum of losses for reference generation and answer generation:
\begin{equation}
\label{eq:loss_rag}
    \mathcal{L}_{\text{rag}} = \mathcal{L}_{\text{ref}} + \mathcal{L}_{\text{ans}}.
\end{equation}
This training approach encourages the CorpusLM not only to generate useful references from documents that will aid in the final answer generation, but also to generate the final answer from the documents and the references. 

\subsection{Unsupervised DocID Understanding Tasks}
Since pre-trained language models lack inherent understanding of DocIDs, we introduce the following tasks to enhance their understanding of DocIDs and align their knowledge with DocIDs:
\begin{itemize}[leftmargin=*]
\item \textbf{Predicting DocIDs from Pseudo Queries:} The model generates a ranked list of relevant DocIDs given a pseudo query, thereby enhancing ranking performance.
\item \textbf{Predicting DocIDs from Document Summaries:} The model predicts the ranking list of DocIDs given a document summary, further improving ranking performance.
\item \textbf{Reciting Content Summary form DocID:} The model generates a summary given a DocID, facilitating better memorization of the main content associated with DocIDs.
\item \textbf{Predicting Related DocIDs form DocID:} The model generates a ranked list of DocIDs related to a given DocID, enabling the model's learning of document-level relevance.
\end{itemize}

For these tasks, pseudo queries and summaries are generated using the Llama2-7B model~\cite{llama2} with specific prompts. To enhance the model's ranking ability, the ranking data augmentation is applied to the above DocID prediction tasks, as discussed in Section~\ref{data_construct}.

All auxiliary tasks are formulated as seq2seq generative tasks. During training, he model is trained to maximize the likelihood of generating the target sequence. Teacher forcing is used to optimize the cross-entropy loss $\mathcal{L}_{\text{aux}}$ in Equation~\eqref{eq:loss_sum}, given by:
\begin{equation}
\label{eq:loss_aux}
    \mathcal{L}_{\text{aux}} = - \sum\nolimits_{i=1}^{T_c} \log f(c_{i}|c_{<i};\theta),
\end{equation}
where $T_c$ represents the length of the target sequence $c$ for the auxiliary task, $c_{i}$ represents the $i$th token in $c$, and $c_{<i}$ corresponds to the generated content up to position $i$.

\begin{table}[t]
\small
\centering
\fontsize{8pt}{\baselineskip}\selectfont
\setlength{\abovecaptionskip}{0.2cm}
\setlength\tabcolsep{4pt}
\caption{Details of datasets in the KILT benchmark.}
\label{tab:datasets}
\begin{tabular}{llrr}
    \toprule
    \textbf{Dataset} & \textbf{Category} & \textbf{Train Size} & \textbf{Dev Size} \\
    \midrule
    FEVER~\cite{fever} & Fact Checking & 104,966 & 10,444 \\
    AIDA CoNLL-YAGO~\cite{aida} & Entity Linking & 18,395 & 4,784 \\
    WNED-WIKI~\cite{wned} & Entity Linking & - & 3,396 \\
    WNED-CWEB~\cite{wned} & Entity Linking & - & 5,599 \\
    T-REx~\cite{trex} & Slot Filling & 2,284,168 & 5,000 \\
    Zero Shot RE~\cite{zsre} & Slot Filling & 147,909 & 3,724 \\
    Natural Questions~\cite{nq} & Open Domain QA & 87,372 & 2,837 \\
    HotpotQA~\cite{hotpotqa} & Open Domain QA & 88,869 & 5,600 \\
    TriviaQA~\cite{triviaqa} & Open Domain QA & 61,844 & 5,359 \\
    ELI5~\cite{eli5} & Open Domain QA & 272,634 & 1,507 \\
    Wizard of Wikipedia~\cite{wow} & Dialogue & 63,734 & 3,054 \\
    \bottomrule
\end{tabular}
\end{table}

\begin{table*}[!ht]
\centering
\setlength{\abovecaptionskip}{0.2cm}
\caption{Overall retrieval performance on KILT dev set. We report passage-level R-Precision (\%). Models are emphasized with the best in \textbf{bold} and the second in \underline{underline}. The symbol "$\dagger$" signifies that our model achieved superior results among all baselines in a statistically significant manner (t-test, $p < 0.05$).}
\label{tab:retrieval}
\setlength\tabcolsep{5.8pt}
\begin{tabular}{lccccccccccc}
    \toprule
    & \textbf{FC} & \multicolumn{3}{c}{\textbf{Entity Linking}} & \multicolumn{2}{c}{\textbf{Slot Filling}} & \multicolumn{4}{c}{\textbf{Open Domain QA}} & \textbf{Dial.} \\
    \cmidrule(lr){2-2}\cmidrule(lr){3-5}\cmidrule(lr){6-7}\cmidrule(lr){8-11}\cmidrule(lr){12-12}
    \multicolumn{1}{l}{\multirow{-2}{*}{\textbf{Method}}} & \textbf{FEVER} & \textbf{AY2} & \textbf{WnWi} & \textbf{WnCw} & \textbf{T-REx} & \textbf{zsRE} & \textbf{NQ} & \textbf{HoPo} & \textbf{TQA} & \textbf{ELI5} & \textbf{WoW} \\
    \midrule
    \multicolumn{12}{l}{\textit{\textbf{Sparse\&Dense Retrieval}}} \\
    BM25 & 30.29 & 2.82 & 1.38 & 3.84 & 32.04 & 43.37 & 12.34 & 31.31 & 14.40 & 1.20 & 17.20 \\
    DPR & 59.10 & 79.51 & - & - & 60.61 & 70.91 & 31.13 & 39.47 & 35.48 & - & 37.66 \\
    MT-DPR & 64.05 & 81.69 & 49.20 & 46.95 & 57.64 & 73.81 & 32.80 & 38.42 & 36.29 & 10.86 & 38.00 \\
    RAG & 66.04 & 76.40 & 48.28 & 46.01 & 53.57 & 67.97 & 38.25 & 34.61 & 41.38 & 10.70 & 38.04 \\
    E5 & 68.52 & 79.72 & 50.47 & 48.10 & 54.48 & 70.01 & 39.40 & 37.35 & 42.62 & \textbf{11.02} & 39.16 \\
    SimLM & 68.06 & 80.11 & 51.98 & 49.54 & 55.42 & 72.11 & 38.58 & 36.11 & 41.80 & 10.36 & 38.31 \\
    \midrule
    \multicolumn{12}{l}{\textit{\textbf{Generative Retrieval}}} \\
    T5 & 71.63 & 86.71 & 67.34 & 62.20 & 64.87 & 78.51 & 38.69 & 38.09 & 45.73 & 10.35 & 42.51 \\
    BART & 69.90 & 87.43 & 67.22 & 60.71 & 61.57 & 76.13 & 39.84 & 38.44 & 47.26 & 10.09 & 40.19 \\
    SEAL & 70.55 & 82.05 & 57.09 & 58.70 & 55.91 & 74.89 & 39.67 & {\ul 40.54} & 44.16 & 9.32 & 41.59 \\
    CorpusBrain & 72.23 & {\ul 88.79} & {\ul 69.40} & 63.23 & 63.42 & 79.05 & 40.09 & 39.45 & 47.97 & 10.68 & 42.19 \\
    Llama2 & 74.39 & 85.53 & 66.55 & 61.45 & 66.12 & 77.90 & 40.59 & 40.37 & 48.43 & 10.66 & 42.69 \\
    \rowcolor[RGB]{236,244,252} 
    CorpusLM (T5) & {\ul 75.64}$^\dagger$ & \textbf{90.96}$^\dagger$ & \textbf{70.35}$^\dagger$ & \textbf{65.43}$^\dagger$ & {\ul 68.89}$^\dagger$ & \textbf{81.08}$^\dagger$ & {\ul 41.46}$^\dagger$ & 39.31 & {\ul 48.80} & {\ul 10.90} & \textbf{44.96}$^\dagger$ \\
    \rowcolor[RGB]{236,244,252} 
    CorpusLM (Llama2) & \textbf{76.21}$^\dagger$ & 88.59 & 69.39 & {\ul 64.18}$^\dagger$ & \textbf{69.17}$^\dagger$ & {\ul 80.79}$^\dagger$ & \textbf{44.10}$^\dagger$ & \textbf{42.06}$^\dagger$ & \textbf{50.62}$^\dagger$ & 10.88 & {\ul 43.92}$^\dagger$ \\
    \bottomrule
\end{tabular}
\end{table*}

\section{Experimental Settings}
\subsection{Datasets}
To evaluate CorpusLM's retrieval and generation performance, we employ KILT (Knowledge-Intensive Language Tasks) benchmark~\cite{kilt}, an extensive benchmark that encompasses 11 datasets across 5 knowledge-intensive natural language processing tasks, including fact checking, entity linking, slot filling, open-domain question answering, and dialogue. We evaluate CorpusLM's performance in retrieval, closed-book generation, and retrieval-augmented generation. Detailed statistics can be found in Table~\ref{tab:datasets}.

The model training is conducted using the training sets, and evaluations are carried out using the development sets. As the knowledge source, we utilize the pre-processed Wikipedia passages split into sections. The passages are derived from English Wikipedia articles based on the 2019/08/01 Wikipedia dump data, consisting of a total of 5.9 million articles and 16.6 million passages. For both dense and sparse models, we refine the passages by adding the title of each article to its corresponding passage, following~\cite{kilt, chen2022corpusbrain}.

\subsection{Evaluation Metrics}
For retrieval evaluation, consistent with previous works~\cite{kilt, genre, seal, mtdpr, rag} and the official KILT evaluation metrics, we use R-precision~\cite{kilt} as the evaluation metric, which is by the fraction $\frac{r}{R}$, with $R$ being the number of contexts in the provenance set, and $r$ is the number of relevant contexts within the top-$R$ retrieved contexts.

For downstream evaluation, we adopt specific metrics for different downstream tasks. Specifically, as suggested in the KILT resource paper~\cite{kilt}, we use Accuracy for FEV, AY2, WnWi, WnCw, T-REx and zsRE; Exact Match (\textit{EM}) for NQ, TQA and HoPo; \textit{ROUGE-L} for ELI5; and \textit{F1} for WoW. To ensure a fair comparison with non-finetuned generators, we include the percentage of outputs containing gold answers ("\textit{has\_answer}"), in line with~\cite{ralle}. 

\subsection{Baselines}
\subsubsection{{Baselines for Retrieval Tasks}}
Retrieval tasks are divided into two main approaches: \textit{Sparse\&Dense Retrieval} and \textit{Generative Retrieval}. 
For \textit{Sparse\&Dense Retrieval}, we employ the following models: \textbf{BM25}~\cite{bm25}, a classic sparse retrieval model; \textbf{DPR}~\cite{dpr} and its multi-task variant \textbf{MT-DPR}~\cite{mtdpr} for dense passage retrieval; \textbf{RAG}~\cite{rag}, which combines dense retrieval with seq2seq models for an enhanced generation; \textbf{E5}~\cite{e5}, a state-of-the-art text embedding model; and \textbf{SimLM}~\cite{simlm}, a dense passage retriever with effective pre-training methods. 
The Generative Retrieval methods comprise \textbf{T5}~\cite{t5}, a pre-trained encoder-decoder model for multi-task learning, \textbf{BART}~\cite{bart}, a denoising autoencoder for text generation; \textbf{SEAL}~\cite{seal}, generating sub-strings as document identifiers; \textbf{CorpusBrain}~\cite{chen2022corpusbrain}, incorporating pre-training strategies for KILT retrieval task; and \textbf{Llama2}~\cite{llama2}, an open-source pre-trained LLM. 

The baseline retrieval models are finetuned with labeled retrieval data from KILT datasets. Except the DPR model is finetuned on each specific dataset, other dense and generative retrieval models are multi-task finetuned using retrieval data across all datasets, as multi-task training in KILT leads to improved performance~\cite{mtdpr, chen2022corpusbrain}.

\subsubsection{{Baselines for Downstream Tasks}}
For downstream tasks, we consider two primary streams: {Closed-book Generation} and {Retrieval-augmented Generation}. 
The Closed-book Generation models include \textbf{T5}~\cite{t5}, \textbf{BART}~\cite{bart}, and the open-source LLM named \textbf{Llama2-70B}~\cite{llama2}, all leveraging their pre-trained parameters for generative tasks. 
For Retrieval-augmented Generation, we consider the following models: (1) \textbf{DPR+BART}~\cite{kilt}, combining dense passage retrieval with generative BART model; (2) \textbf{RAG}~\cite{rag}, an end-to-end retriever-generator model; (3) \textbf{MT-DPR+FID}~\cite{mtdpr}, integrating multi-task DPR retriever with a fusion-in-decoder generator~\cite{fid}; (4) \textbf{BM25+Llama2-70B} and \textbf{E5+Llama2-13B\&70B} models reported by~\cite{ralle} that utilize Llama2~\cite{llama2} for answer generation.

\begin{table*}[!ht]
\centering
\setlength{\abovecaptionskip}{0.2cm}
\caption{Overall downstream performance on KILT dev set, considering both closed-book and RAG settings. 
In different settings, models are emphasized with the best in \textbf{bold} and the second in \underline{underline}. $^\diamondsuit$ and $^\heartsuit$ indicates results from~\cite{kilt} and~\cite{ralle}, respectively. 
Following~\cite{ralle}, the percentage of outputs containing gold answers ("\textit{has\_answer}") is noted in parentheses.}
\label{tab:generation}
\setlength\tabcolsep{4.8pt}
\renewcommand{\arraystretch}{1.03} 
\begin{tabular}{lccccccccccc}
    \toprule
    & \textbf{FC} & \multicolumn{3}{c}{\textbf{Entity Linking}} & \multicolumn{2}{c}{\textbf{Slot Filling}} & \multicolumn{4}{c}{\textbf{Open Domain QA}} & \textbf{Dial.} \\
    \cmidrule(lr){2-2}\cmidrule(lr){3-5}\cmidrule(lr){6-7}\cmidrule(lr){8-11}\cmidrule(lr){12-12}
    \multirow{-2}{*}{\textbf{Method}} & \textbf{FEVER} & \textbf{AY2} & \textbf{WnWi} & \textbf{WnCw} & \textbf{T-REx} & \textbf{zsRE} & \textbf{NQ} & \textbf{HoPo} & \textbf{TQA} & \textbf{ELI5} & \textbf{WoW} \\
    \midrule
    \multicolumn{12}{l}{\textit{\textbf{Closed-book Generation}}} \\
    T5{$^\diamondsuit$} & - & 81.84 & 47.35 & 46.58 & 47.24 & 1.58 & 25.20 & 12.66 & 25.79 & 21.02 & 13.15 \\
    BART{$^\heartsuit$} & 80.67 & 86.62 & 47.91 & 48.01 & 43.84 & 3.03 & 26.15 & 16.86 & 32.54 & 22.69 & 13.77 \\
    Llama2-70B{$^\heartsuit$} & 33.6\footnotesize{(74.9)}\normalsize & 39.8\footnotesize{(54.5)}\normalsize & 42.8\footnotesize{(53.8)}\normalsize & 39.2\footnotesize{(55.7)}\normalsize & 28.5\footnotesize{(40.5)}\normalsize & {\ul 11.3\footnotesize{(13.6)}\normalsize} & 19.6\footnotesize{(37.4)}\normalsize & 13.9\footnotesize{(25.1)}\normalsize & \textbf{67.4\footnotesize{(80.8)}\normalsize} & \textbf{23.00} & 13.30 \\
    \rowcolor[RGB]{236,244,252} 
    CorpusLM (T5) & {\ul 81.93} & \textbf{87.16} & {\ul 49.68} & \textbf{51.65} & {\ul 49.65} & 3.61 & {\ul 28.14} & {\ul 17.32} & 25.02 & 21.33 & {\ul 13.90} \\
    \rowcolor[RGB]{236,244,252} 
    CorpusLM (Llama2) & \textbf{85.34} & {\ul 86.91} & \textbf{56.66} & {\ul 50.37} & \textbf{51.05} & \textbf{18.22} & \textbf{29.57} & \textbf{26.13} & {\ul 41.34} & {\ul 22.94} & \textbf{14.85} \\
    \midrule
    \multicolumn{12}{l}{\textit{\textbf{Retrieval-augmented Generation}}} \\
    DPR+BART{$^\diamondsuit$} & 88.11 & - & 44.96 & 45.70 & 56.70 & 34.96 & 45.05 & 25.75 & 59.28 & 18.53 & 15.51 \\
    RAG{$^\diamondsuit$} & 87.70 & 77.40 & 49.00 & 46.70 & 61.48 & 47.42 & 48.78 & 27.68 & 61.73 & 16.11 & 13.28 \\
    MT-DPR+FID & 88.49 & 79.77 & 49.52 & 47.15 & 79.43 & 69.09 & 50.07 & 36.50 & 69.62 & 15.77 & 15.60 \\
    BM25+Llama2-70B{$^\heartsuit$} & 46.2\footnotesize{(86.3)}\normalsize & 18.0\footnotesize{(35.9)}\normalsize & 19.1\footnotesize{(32.2)}\normalsize & 14.2\footnotesize{(30.9)}\normalsize & 25.9\footnotesize{(43.0)}\normalsize & 31.4\footnotesize{(37.8)}\normalsize & 25.3\footnotesize{(34.3)}\normalsize & 25.9\footnotesize{(33.4)}\normalsize & 65.8\footnotesize{(80.0)}\normalsize & 21.30 & 12.20 \\
    E5+Llama2-13B{$^\heartsuit$} & 66.3\footnotesize{(73.5)}\normalsize & 51.2\footnotesize{(57.9)}\normalsize & 48.6\footnotesize{(51.4)}\normalsize & 45.6\footnotesize{(51.4)}\normalsize & 17.2\footnotesize{(42.3)}\normalsize & 31.7\footnotesize{(41.1)}\normalsize & 36.1\footnotesize{(43.3)}\normalsize & 14.3\footnotesize{(25.5)}\normalsize & 56.3\footnotesize{(76.2)}\normalsize & 20.90 & 12.30 \\
    E5+Llama2-70B{$^\heartsuit$} & 49.9\footnotesize{(88.6)}\normalsize & 51.2\footnotesize{(57.9)}\normalsize & 48.6\footnotesize{(51.4)}\normalsize & 45.6\footnotesize{(51.4)}\normalsize & 28.9\footnotesize{(49.2)}\normalsize & 35.0\footnotesize{(43.2)}\normalsize & 36.4\footnotesize{(48.8)}\normalsize & 28.1\footnotesize{(35.8)}\normalsize & {\ul 71.1\footnotesize{(83.9)}\normalsize} & 21.50 & 13.20 \\
    \rowcolor[RGB]{236,244,252} 
    CorpusLM (T5) & {\ul 89.81} & \textbf{87.09} & {\ul 50.52} & {\ul 49.77} & {\ul 80.68} & {\ul 70.34} & \textbf{53.39} & {\ul 40.96} & 70.94 & {\ul 22.13} & {\ul 16.65} \\
    \rowcolor[RGB]{236,244,252} 
    CorpusLM (Llama2) & \textbf{90.22} & {\ul 85.03} & \textbf{56.54} & \textbf{50.32} & \textbf{81.57} & \textbf{72.79} & {\ul 55.38} & \textbf{42.23} & \textbf{72.43} & \textbf{23.46} & \textbf{16.96} \\
    \bottomrule
\end{tabular}
\end{table*}

\subsection{Implementation Details}
In our experiments, we utilize the T5-Base~\cite{t5} (T5) and Llama2-7B-Chat~\cite{llama2} (Llama2) as our backbone models with pre-trained parameters from Hugging Face~\cite{wolf2019huggingface}. 

During training, we set the coefficients $\lambda_{1-4}$ to 1 and the noise sampling probability $\tau$ to 0.2. Our training process involves a batch size of 512 and a learning rate of 3e-4. We use 2000 warm-up steps for the learning rate. To improve the training efficiency of the CorpusLM (Llama2) model, we employ QLoRA~\cite{qlora} and DeepSpeed~\cite{deepspeed, zero} technologies. We incorporate a maximum of top-10 and top-3 ranked passages for training the DocID decoding and answer decoding, respectively. Each unsupervised DocID understanding task consists of approximately 300k training pairs.

During inference, we use dynamic constrained greedy decoding for generative retrieval tasks and greedy decoding for downstream tasks. For retrieval tasks, we limit the number of generated DocIDs to a maximum of 10. For RAG tasks, we retrieve and utilize 3 passages as the contexts for downstream tasks. All experiments are conducted on 8 NVIDIA Tesla A100 40GB GPUs.

\section{Experimental Results}
\subsection{Retrieval Performance}
In this section, we evaluate the retrieval performance of our proposed CorpusLM models based on T5 and LLama2, compared with a range of baseline models, as presented in Table~\ref{tab:retrieval}. 

\textbf{Comparison with Sparse\&Dense Retrievers:}
Despite the competitive performance of dense retrievers like MT-DPR, E5, RAG, and SimLM, which have been fine-tuned through multi-task training, \textbf{our CorpusLM models achieve superior performance in most datasets by a significant margin (t-test, $p < 0.05$)}. Specifically, on the FEVER dataset, T5 and LLama2-based CorpusLM outperform MT-DPR by 18.10\% and 18.99\% respectively. Similarly, on the zsRE dataset, CorpusLM (T5) and CorpusLM (LLama2) surpass SimLM by 12.44\% and 12.03\% respectively.
We also observe that the bag-of-words method BM25, a sparse retrieval model, falls behind neural retrieval models in all tasks, highlighting the need for neural models due to the complexity of these tasks. Notably, joint training of the DPR model on multiple datasets (MT-DPR) outperforms individual dataset training in almost all cases, indicating the effectiveness of joint training for improved retrieval performance.

\textbf{Comparison with Generative Retrievers:}
Generative retrievers generally outperform dense retrievers, indicating their effectiveness in knowledge-intensive tasks. However, they still fall short compared to our CorpusLM models. T5-based CorpusLM and LLama2-based CorpusLM models generally exhibit similar retrieval performance. LLama2-based CorpusLM performs better in fact-checking and open domain QA tasks, while T5-based CorpusLM outperforms in entity linking and dialogue tasks.
\textbf{Compared to other baseline generative retrievers, our CorpusLM models consistently achieve superior performance across most datasets.} For example, on the WoW dataset, CorpusLM (T5) and CorpusLM (LLama2) outperform T5 by 5.76\% and 3.31\% respectively, and surpass CorpusBrain by 6.57\% and 4.11\% respectively. On the T-REx dataset, CorpusLM (T5) and CorpusLM (LLama2) outstrip SEAL by 23.21\% and 23.72\%, and surpass LLama2 by 4.19\% and 4.61\% respectively.


\begin{table*}[!ht]
\centering
\setlength{\abovecaptionskip}{0.2cm}
\caption{Ablation studies on retrieval performance. The best results are in bold for T5 and Llama2-based models, respectively.}
\label{tab:ablation_retr}
\setlength\tabcolsep{5pt}
\renewcommand{\arraystretch}{1.02} 
\begin{tabular}{lccccccccccc}
    \toprule
    & \textbf{FC} & \multicolumn{3}{c}{\textbf{Entity Linking}} & \multicolumn{2}{c}{\textbf{Slot Filling}} & \multicolumn{4}{c}{\textbf{Open Domain QA}} & \textbf{Dial.} \\
    \cmidrule(lr){2-2}\cmidrule(lr){3-5}\cmidrule(lr){6-7}\cmidrule(lr){8-11}\cmidrule(lr){12-12}
    \multicolumn{1}{l}{\multirow{-2}{*}{\textbf{Method}}} & \textbf{FEVER} & \textbf{AY2} & \textbf{WnWi} & \textbf{WnCw} & \textbf{T-REx} & \textbf{zsRE} & \textbf{NQ} & \textbf{HoPo} & \textbf{TQA} & \textbf{ELI5} & \textbf{WoW} \\
    \midrule
    \rowcolor[RGB]{236,244,252} 
    \textbf{CorpusLM (T5)} & \textbf{75.64} & \textbf{90.96} & 70.35 & 65.43 & \textbf{68.89} & \textbf{81.08} & \textbf{41.46} & \textbf{39.31} & \textbf{48.80} & \textbf{10.90} & \textbf{44.96} \\
    \hspace{1em}w/o DocID Understanding Tasks & 74.25 & 89.23 & 67.32 & 63.02 & 68.45 & 80.30 & 39.23 & 37.24 & 46.28 & 8.89 & 42.75 \\
    \hspace{1em}w/o Ranking-oriented Decode & 74.03 & 90.67 & \textbf{70.65} & \textbf{65.97} & 67.98 & 80.78 & 40.17 & 37.13 & 46.92 & 10.06 & 44.70 \\
    \midrule
    \rowcolor[RGB]{236,244,252} 
    \textbf{CorpusLM (Llama2)} & \textbf{76.21} & \textbf{88.59} & \textbf{69.39} & \textbf{64.18} & \textbf{69.17} & \textbf{80.79} & \textbf{44.10} & \textbf{42.06} & \textbf{50.62} & \textbf{10.88} & \textbf{43.92} \\
    \hspace{1em}w/o DocID Understanding Tasks & 75.28 & 87.58 & 66.13 & 62.75 & 69.07 & 79.37 & 42.83 & 40.82 & 49.24 & 10.08 & 42.59 \\
    \hspace{1em}w/o Ranking-oriented Decode & 72.79 & 84.20 & 64.50 & 60.95 & 66.59 & 75.59 & 41.20 & 39.20 & 47.50 & 9.10 & 40.90 \\
    \bottomrule
\end{tabular}
\end{table*}

\begin{table*}[!ht]
\centering
\setlength{\abovecaptionskip}{0.2cm}
\caption{Ablation studies on downstream performance in RAG setting. The best results are in bold for T5 and Llama2-based models, respectively.}
\label{tab:ablation_rag}
\setlength\tabcolsep{6pt}
\renewcommand{\arraystretch}{1.02} 
\begin{tabular}{p{3.6cm}ccccccccccc}
    \toprule
    & \textbf{FC} & \multicolumn{3}{c}{\textbf{Entity Linking}} & \multicolumn{2}{c}{\textbf{Slot Filling}} & \multicolumn{4}{c}{\textbf{Open Domain QA}} & \textbf{Dial.} \\
    \cmidrule(lr){2-2}\cmidrule(lr){3-5}\cmidrule(lr){6-7}\cmidrule(lr){8-11}\cmidrule(lr){12-12}
    \multicolumn{1}{l}{\multirow{-2}{*}{\textbf{Method}}} & \textbf{FEVER} & \textbf{AY2} & \textbf{WnWi} & \textbf{WnCw} & \textbf{T-REx} & \textbf{zsRE} & \textbf{NQ} & \textbf{HoPo} & \textbf{TQA} & \textbf{ELI5} & \textbf{WoW} \\
    \midrule
    \rowcolor[RGB]{236,244,252} 
    \textbf{CorpusLM (T5)} & \textbf{89.81} & \textbf{87.09} & \textbf{50.52} & \textbf{49.77} & \textbf{80.68} & \textbf{70.34} & \textbf{53.39} & \textbf{40.96} & \textbf{70.94} & 22.13 & \textbf{16.65} \\
    \hspace{1em}w/o Decode Reference & 87.21 & 84.17 & 49.09 & 47.97 & 77.95 & 67.92 & 50.47 & 36.42 & 68.50 & 21.16 & 15.85 \\
    \hspace{1em}w/o Noise Sampling & 88.31 & 86.48 & 49.41 & 48.53 & 79.14 & 68.90 & 52.78 & 38.81 & 70.29 & 20.17 & 15.05 \\
    \hspace{1em}w/o Pipeline Decode & 89.18 & 86.12 & 48.38 & 47.07 & 78.51 & 67.83 & 52.56 & 39.06 & 69.82 & \textbf{22.39} & 16.20 \\
    \midrule
    \rowcolor[RGB]{236,244,252} 
    \textbf{CorpusLM (Llama2)} & \textbf{90.22} & 85.03 & \textbf{56.54} & \textbf{50.32} & \textbf{81.57} & \textbf{72.79} & \textbf{55.38} & \textbf{42.23} & \textbf{72.43} & \textbf{23.46} & \textbf{16.96} \\
    \hspace{1em}w/o Decode Reference & 87.40 & 84.49 & 55.14 & 47.89 & 79.61 & 70.53 & 51.88 & 38.25 & 69.58 & 21.61 & 15.70 \\
    \hspace{1em}w/o Noise Sampling & 88.04 & \textbf{85.43} & 55.42 & 48.65 & 79.77 & 71.51 & 54.25 & 39.66 & 71.79 & 21.31 & 15.66 \\
    \bottomrule
\end{tabular}
\end{table*}

\subsection{Downstream Generation Performance}

In this section, we analyze the downstream generation performance of the models within the closed-book and open-retrieval settings on the KILT dev set. The results are shown in Table~\ref{tab:generation}.

In the closed-book setting, \textbf{our CorpusLM models surpass traditional language models like T5 and BART, and even outperform the larger model Llama2-70B, across most datasets.} While both CorpusLM variations demonstrate similar performance, the larger parameter scale of CorpusLM (Llama2) provides a slight edge over CorpusLM (T5) across most datasets.
The T5-based CorpusLM notably outshines T5 model by 10.88\%, 11.67\%, and 36.81\% on the WnCw, NQ, and HoPo datasets respectively, and even surpasses the Llama2-70B by 31.76\%, 43.57\%, and 24.60\%.
Furthermore, on the FEVER and T-REx datasets, the Llama2-7B-based CorpusLM surpasses the Llama2-70B by 153.99\% and 79.12\%, respectively. They also excel in the \textit{has\_answer} metric by 13.94\% and 26.05\%. 
These results demonstrate the effectiveness of our multi-task learning framework for closed-book generation tasks.

In the RAG setting, \textbf{our CorpusLM models also take the lead, with CorpusLM (Llama2) demonstrating the most substantial improvements.} 
Among small language models, our T5-based CorpusLM outperforms RAG by 12.52\%, 47.98\%, and 37.37\% on the AY2, HoPo, and ELI5 datasets respectively. It even exceeds MT-DPR+FID, which utilizes a stronger fusion-in-decoder generator, by 9.17\%, 12.21\%, and 40.36\% on these datasets.
Among LLMs, our Llama2-based CorpusLM surpasses E5+Llama2-13B by 36.08\%, 129.64\%, and 37.92\% on the FEVER, zsRE, and WoW datasets respectively. It even outperforms E5+Llama2-70B by 80.80\%, 107.99\%, and 28.51\% on these datasets. 
The above results validate the effectiveness of our multi-task learning framework and continuous RAG strategy.

\begin{figure*}[!t]
    \centering
    \setlength{\abovecaptionskip}{0.2cm}
    \includegraphics[width=0.875\linewidth]{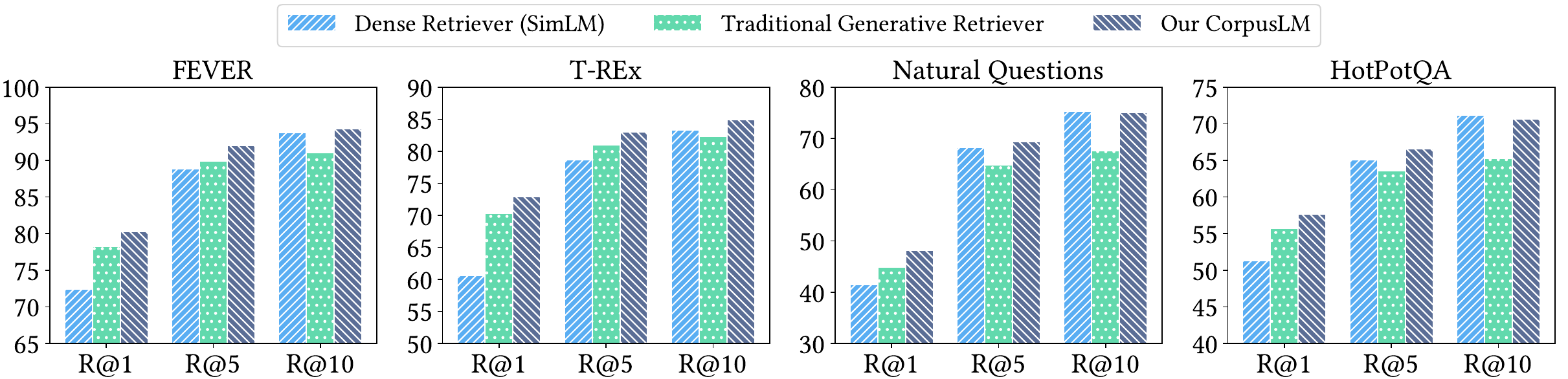}
    \caption{
    Analysis of ranking capability of CorpusLM for retrieval tasks. We compare our T5-based CorpusLM with dense retriever SimLM and another T5-based generative retriever with traditional beam search, focusing on Recall@\{1, 5, 10\}.
    }%
    \label{fig:recall_bars}
\end{figure*}

\subsection{Ablation Studies}
In this section, we conduct experiments on ablations for both retrieval and downstream tasks, to assess the effectiveness of each proposed component for T5 and Llama2-based CorpusLM.

\subsubsection{Ablations for Retrieval Tasks}
For retrieval tasks, we investigate the impact of the DocID understanding tasks and the DocID list decoding strategy. The results are shown in Table \ref{tab:ablation_retr}.

\textbf{(1) Removing DocID understanding tasks reduces retrieval performance across all datasets for both T5 and LLama2-based CorpusLM}. Specifically, we see notable decreases on the NQ, HoPo, and ELI5 datasets by $5.37\%$, $5.53\%$, and $18.42\%$ for CorpusLM (T5), respectively. For CorpusLM (Llama2), we also see notable declines on the WnWi and ELI5 datasets, with drops of $4.70\%$ and $7.40\%$, respectively. This underlines the crucial role of DocID understanding in enhancing CorpusLM's retrieval capability.
\textbf{(2) Changing the ranking-oriented decoding to the traditional beam decoding leads to lower R-Precision in most datasets}. Notably, for CorpusLM (T5), TQA and ELI5 saw reductions of $3.84\%$ and $7.74\%$ in R-Precision, highlighting the importance of this strategy for retrieval. For CorpusLM (Llama2), we also observe a consistent performance drop without the DocID understanding tasks, 

\subsubsection{Ablations for Downstream Tasks}
We further examine the effectiveness of the proposed reference decoding, noise sampling, and continuous decoding. The results are detailed in Table~\ref{tab:ablation_rag}.

\textbf{(1) Removing reference decoding, there's a performance decline across all tasks.} This is particularly pronounced on NQ and HoPo datasets, with a decrease of $2.92\%$ and $4.54\%$, respectively. For CorpusLM (Llama2), removing reference decoding notably affects the accuracy in the Open Domain QA domain, especially for TQA and ELI5, with reductions of $2.85\%$ and $1.85\%$, respectively. This indicates that reference decoding plays a critical role in filtering out irrelevant information and generating more accurate answers.
%
\textbf{(2) Without noise sampling, we also observe a reduction in most datasets}, with decreases of $0.61\%$ and $1.44\%$ on AY2 and on zsRE for T5-based CorpusLM, respectively. This suggests that introducing noise during training helps improve RAG performance.
\textbf{(3) The removal of continuous decoding leads to drops in most datasets for CorpusLM (T5)}, with a decline of $2.14\%$ on WnWi and $1.12\%$ on TQA, showcasing the importance of a streamlined decoding process.

\subsection{Analysis of Ranking Capability}
To evaluate the effectiveness of the proposed ranking-oriented DocID decoding strategy, we conduct a series of experiments to compare the dense retriever (DR) SimLM~\cite{simlm}, generative retriever (GR), and our T5-based CorpusLM, with a focus on the Recall metric. Detailed comparisons are shown in Figure~\ref{fig:recall_bars}, where the generative retriever utilizes our CorpusLM model with the traditional constrained beam decoding method. 

For the four datasets in the figure, we observe that Dense Retriever is consistently outperformed by Generative Retriever for Recall@1, but it surpasses GR at Recall@10. Notably, on the FEVER and NQ datasets, DR lags behind GR by 7.41\% and 7.81\% at Recall@1, yet it leads by 3.01\% and 11.44\% at Recall@10, respectively. This trend suggests that traditional GR methods are firmly focused on the top-1 ranked DocID, often at the expense of accuracy for other rop-ranked DocIDs. 
\textbf{Our CorpusLM, however, demonstrates superior performance to GR at all levels of Recall}. With improvements of 3.68\% and -10.94\% in Recall@10 for FEVER and NQ, our model showcases an advanced capability for ranking enhancement. Moreover, CorpusLM consistently outpaces the DR at Recall@5 and is competitive at Recall@10.
These results underline the strength of the ranking strategy employed by CorpusLM, leading to improved ranking performance.

\begin{table}[t]
\centering
\fontsize{8.5pt}{\baselineskip}\selectfont
\setlength{\abovecaptionskip}{0.2cm}
\setlength\tabcolsep{4.5pt}
\caption{Analysis of the efficiency of RAG models.}
\label{tab:efficiency}
\begin{tabular}{lc*{2}{>{\centering\arraybackslash}p{1.45cm}}}
    \toprule
    \textbf{Method} & \textbf{Parameters} & \textbf{Storage} & \textbf{Latency} \\ 
    \midrule
    RAG & 626M & 59.3G & 106.7ms \\
    MT-DPR+FID & 440M & 51.2G & 160.9ms \\
    \rowcolor[RGB]{236,244,252} 
    \textbf{CorpusLM (T5)} & \textbf{220M} & \textbf{426.1M} & \textbf{78.4ms} \\
    \hspace{0.5em}w/ Pipeline Decode & 220M & 426.1M & 102.9ms \\
    \bottomrule
\end{tabular}
\end{table}

\subsection{Efficiency Analysis for RAG Tasks}

We then analyze the efficiency of our continuous RAG strategy, evaluating model parameters, memory footprint and query latency, as shown in Table~\ref{tab:efficiency}.
Our CorpusLM (T5) showcases a significant reduction in the number of parameters, memory usage, and query latency compared to other methods such as RAG and MT-DPR+FID. Specifically, the CorpusLM model has approximately \textbf{2.8-fold fewer parameters} than the RAG~\cite{rag} model, resulting in a substantially lower computational cost. Notably, it implies a \textbf{139.1-fold decrease in memory footprint}, primarily due to the model only requiring storage for titles and sections as DocIDs rather than a large-scale dense document index.
Moreover, compared with MT-DPR+FID~\cite{mtdpr}, the CorpusLM model exhibits a \textbf{2.1-fold reduction in query latency}. The table also highlights that even when utilizing a pipeline decode approach, our model maintains its efficiency advantage, with a 1.3-fold faster decoding speed compared to the multi-round input-output RAG method.
Our CorpusLM model utilizes continuous decoding to combine generative retrieval and answer generation into a single step, which can accelerate response times and cut down on the computational costs of repeated interaction cycles in traditional RAG approaches.

\section{Conclusion and Future Works}
In this work, we propose CorpusLM, a unified language model that handles various knowledge-intensive tasks by integrating generative retrieval, closed-book generation, and RAG. By unifying all tasks and incorporating auxiliary DocID understanding tasks, we facilitate joint learning of retrieval and generation, thus enhance the model's performance in KI tasks. 
This work opens up the potential for unifying various IR tasks (especially the retrieval task that cannot be easily performed by LLMs) into a single generative language model. 
Furthermore, our model is enhanced with two well-designed generation strategies: (1) a ranking-oriented DocID list generation strategy, which refines GR by directly learning from a DocID ranking list, to improve retrieval quality. (2) a continuous DocIDs-References-Answer generation strategy, which facilitates effective and efficient RAG. These generation strategies enable the model to perform retrieval and RAG in a continuous auto-regressive decoding process. 
Experiments conducted on 11 datasets in the KILT benchmark have shown our CorpusLM's superior performance in both retrieval and downstream tasks. 


\begin{acks}
Zhicheng Dou is the corresponding author. This work was supported by Beijing Natural Science Foundation No. L233008, National Natural Science Foundation of China No. 62272467,  the fund for building world-class universities (disciplines) of Renmin University of China, and Public Computing Cloud, Renmin University of China. The work was partially done at the Engineering Research Center of Next-Generation Intelligent Search and Recommendation, MOE, and Beijing Key Laboratory of Big Data Management and Analysis Methods.
\end{acks}

\clearpage
\bibliographystyle{ACM-Reference-Format}
\bibliography{main}

\end{document}